\documentclass[runningheads]{llncs}
\usepackage{amsmath}
\usepackage{graphicx}
\usepackage{subcaption}
\usepackage{listings}
\usepackage{float}
\begin{document}

\title{Geo-locating Road Objects using Inverse Haversine Formula and NVIDIA DriveWorks}
\titlerunning{Geo-locating Road Objects}
%
\author{Mamoona Birkhez Shami\inst{1}\orcidID{0000-0001-8754-3259} \and
Gabriel Kiss\inst{1} \and
Trond Arve Haakonsen\inst{2} \and
Frank Lindseth\inst{1}}
\authorrunning{M. B. Shami et al.}
%
\institute{Department of Computer Science, Norwegian University of Science and Technology (NTNU), Norway \and
Norwegian Public Roads Administration (NPRA), Norway}
\maketitle              
\begin{abstract}
Geolocation is integral to the seamless functioning of autonomous vehicles and advanced traffic monitoring infrastructures. This paper introduces a methodology to geolocate road objects using a monocular camera, leveraging the NVIDIA DriveWorks platform. We use the Centimeter Positioning Service (CPOS) and the inverse Haversine formula to geo-locate road objects accurately. The real-time algorithm processing capability of the NVIDIA DriveWorks platform enables instantaneous object recognition and spatial localization for Advanced Driver Assistance Systems (ADAS) and autonomous driving platforms. We present a measurement pipeline suitable for autonomous driving (AD) platforms and provide detailed guidelines for calibrating cameras using NVIDIA DriveWorks. Experiments were carried out to validate the accuracy of the proposed method for geolocating targets in both controlled and dynamic settings. We show that our approach can locate targets with less than 1m error when the AD platform is stationary and less than 4m error at higher speeds (i.e. up to 60km/h) within a 15m radius.  


\end{abstract}

\section{Introduction}
Autonomous driving and intelligent transportation systems have become a focal point of interest in academia and industry over recent years \cite{yurtsever2020survey,grigorescu2020survey}. Digital Twins can help in traffic management and optimize transportation planning. Information from intelligent vehicles can be streamed to the digital twins to update the virtual twin. The digital twins of urban areas can potentially save maintenance costs and time by enabling efficient supervision and predicting maintenance demands. At the same time, they can reduce the rate of accidents and save precious lives \cite{wu2021digital}. At the core of these systems lies the capacity to perceive the environment and understand the relative positioning of objects. Geolocation, or the ability to determine the precise location of objects on the Earth's surface, plays a pivotal role in this endeavour. Therefore, developing appropriate algorithms for geolocation is one of the first and vital steps in developing these systems.

Given their low cost and widespread use in vehicles, monocular cameras offer an attractive modality for infrastructure perception. However, inferring depth and geolocation information from a single viewpoint is a complex problem. Therefore, traditionally, many solutions rely on support from multi-sensor fusion or stereoscopic cameras. Relying on multiple sensors introduces complexity in the system, which is prone to error and is difficult to maintain. On the other hand, coupling this system only with the commonly available global position system (GPS) without multiple sensors often leads to sub-optimal results due to errors in the positioning system. Global navigation satellite system (GNSS) with CPOS provides more accurate results, but most GNSS accuracy parameters are given for a perfectly stationary scenario, which is not our case.  

In this paper, we address the geolocation of road objects using a monocular camera setup, harnessing the computational capabilities of the NVIDIA DriveWorks platform. Initially designed for real-time object recognition and image processing, this platform provides a suitable framework to integrate and execute our proposed methodologies. NVIDIA Driveworks and Openpilot are among the most used platforms for AD in the real world \cite{yurtsever2020survey}. We provide critical guidelines needed to calibrate the NVIDIA DriveWorks successfully. We utilize the inverse Haversine formula to geolocate the road objects. The focus of our method is not on the object detection pipeline, which is the focus of many previous studies \cite{6130242,SOHEILIAN20131}. Instead, we focus on proposing a practical method of determining the position of any given object in real-time with high accuracy. We test the effectiveness of our approach in controlled and dynamic scenarios, demonstrating that it maintains a low error in all conditions. 

Our major contributions are stated as follows:
\begin{itemize}

\item We propose an inverse Haversine formula-based algorithm to geolocate road objects accurately in car-mounted camera images.

\item  We demonstrate the effectiveness of the geolocation algorithm on control markers and traffic signs with stationary and moving vehicles. We test geolocalization in different scenarios and at different speeds and distances. 

\item We use NVIDIA DriveWorks to implement our pipeline and to calibrate the cameras of the AD platform. We share best practices to calibrate the system. To the best of our knowledge, no other independent work validates the calibration accuracy done by NVIDIA DriveWorks.

\end{itemize}

\section{Related Works}


Researchers have presented various methods for geolocating targets. Timofte et al.~\cite{6130242} and Soheilian et al.~\cite{SOHEILIAN20131} carried out seminal work in 3D localization for road fixtures and traffic signs by using multi-view images. Much Work has been done for geolocating objects using a monocular camera. Oosterman et al.\cite{oosterman2010geolocation} detect signs but geolocate using a GoogleMap geocoder. Some works propose crowd-sourcing images but rely on Google API for geolocation \cite{novais2017community}. Hebbalaguppe et al.~\cite{7926669} employed object detection on Google Street View images for telecom inventory. They used image triangulation for localization, but they made a lot of assumptions about the images and did not report the measurement error. Kuutti et al.\cite{kuutti2018survey} discuss many state-of-the-art localization methods for AD platforms. The closest work to our proposed method can be the one proposed by Namazi et al.\cite{namazi2022geolocation}. They use a low-cost GPS sensor and image processing to localize moving targets, but they do not use NVIDIA Driveworks or report the reliability of their ground truth and how their measurements were verified. On the other hand, we collected our ground truth and focused on the precise calibration of cameras to ensure our measurements were reliable and accurate.

\section{Method}
\begin{figure}
    \centering
    \includegraphics[scale=0.4]{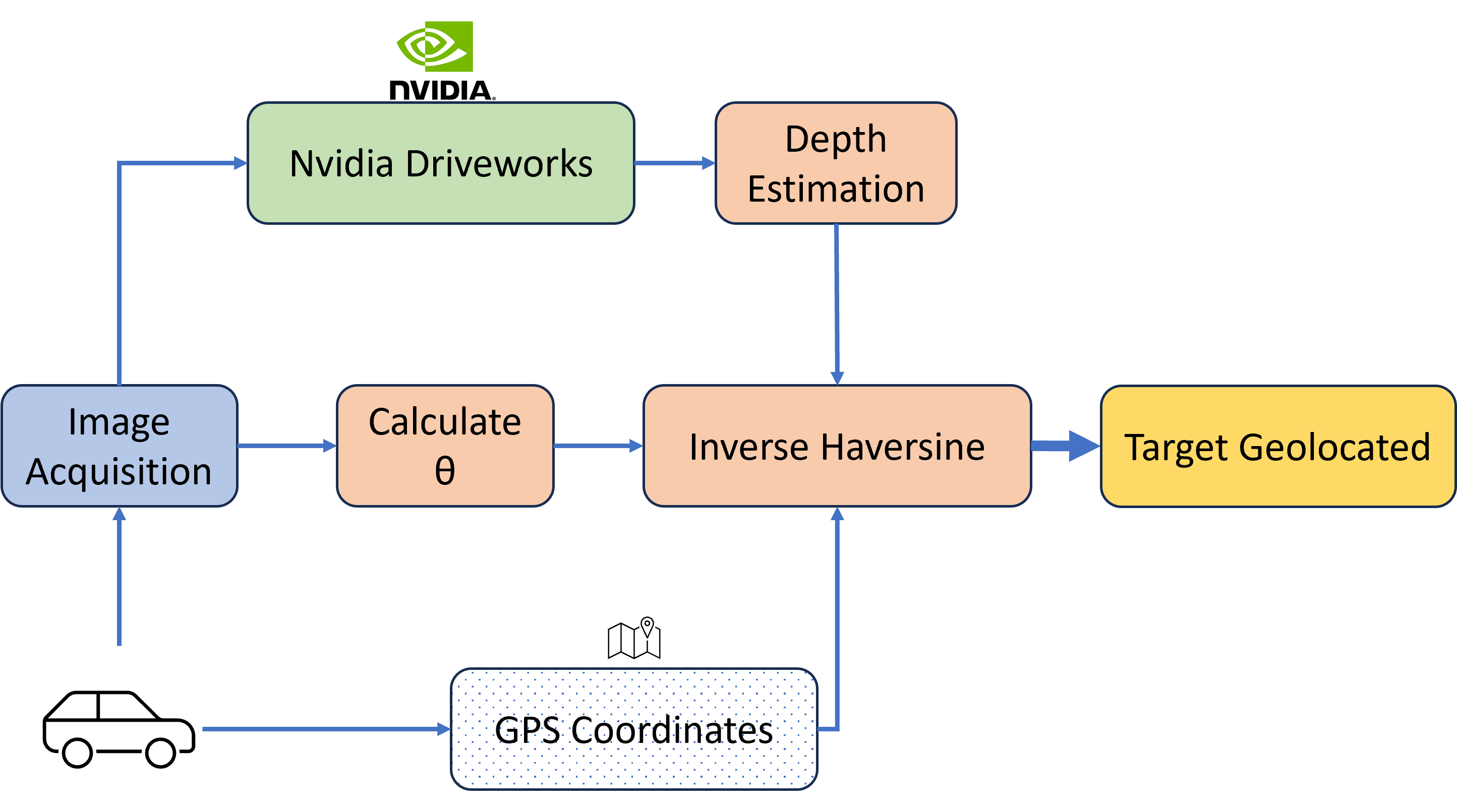}
    \caption{Flowchart for the proposed method of geolocation.}
    \label{fig:process}
\end{figure}

We propose to leverage NVIDIA DriveWorks for depth estimation and use the inverse haversine formula to find the position of the targets given the position of the car as input. The complete pipeline of the proposed method is shown in Fig. \ref{fig:process}. Images are acquired from the cameras mounted on the car. The target is manually annotated in the images. The target's distance from the vehicle is estimated with the help of NVIDIA DriveWorks. The heading of the target is acquired through image processing. The vehicle's position, target's distance and heading are then fed to the inverse haversine formula to geolocate the target. The target may be either a control marker or the traffic signs in the test area.

Experimental and validation data for this work are acquired from different sources. The details of data sources are given in section \ref{sec:data}. The accuracy of the target's estimated position depends on the target's estimated distance from the vehicle. We implemented a custom app that uses NVIDIA DriveWorks and calculates the distance to any given point located in the ground plane in a camera frame. This estimation requires that the cameras are calibrated accurately using NVIDIA's guidelines. The calibration method is also discussed in section \ref{sec:calib}. The formulation and calculations for the input parameters of the inverse Haversine formula are detailed in section \ref{sec:geoest}.

\subsection{Data}
\label{sec:data}

The experiments were carried out using the autonomous vehicle from the NTNU's Autonomous Perception Lab (NAPLab). The ground truth for different targets was acquired manually using the centimetre-accurate CPOS service. The National Road Database (NVDB) was also considered as a ground truth source for traffic signs because it is publicly available data that is easy to access. However, it was decided not to use the data from NVDB as ground truth due to significant errors found in the positions of some traffic signs and the unreliability of its accuracy. Still, the method to access data from the NVDB API is discussed here since NVDB may serve as a reliable ground truth in future works.

\subsubsection{Data Acquisition with Autonomous Vehicle} \label{sec:cardaq}

\begin{figure}
    \centering
    \includegraphics[scale=0.3]{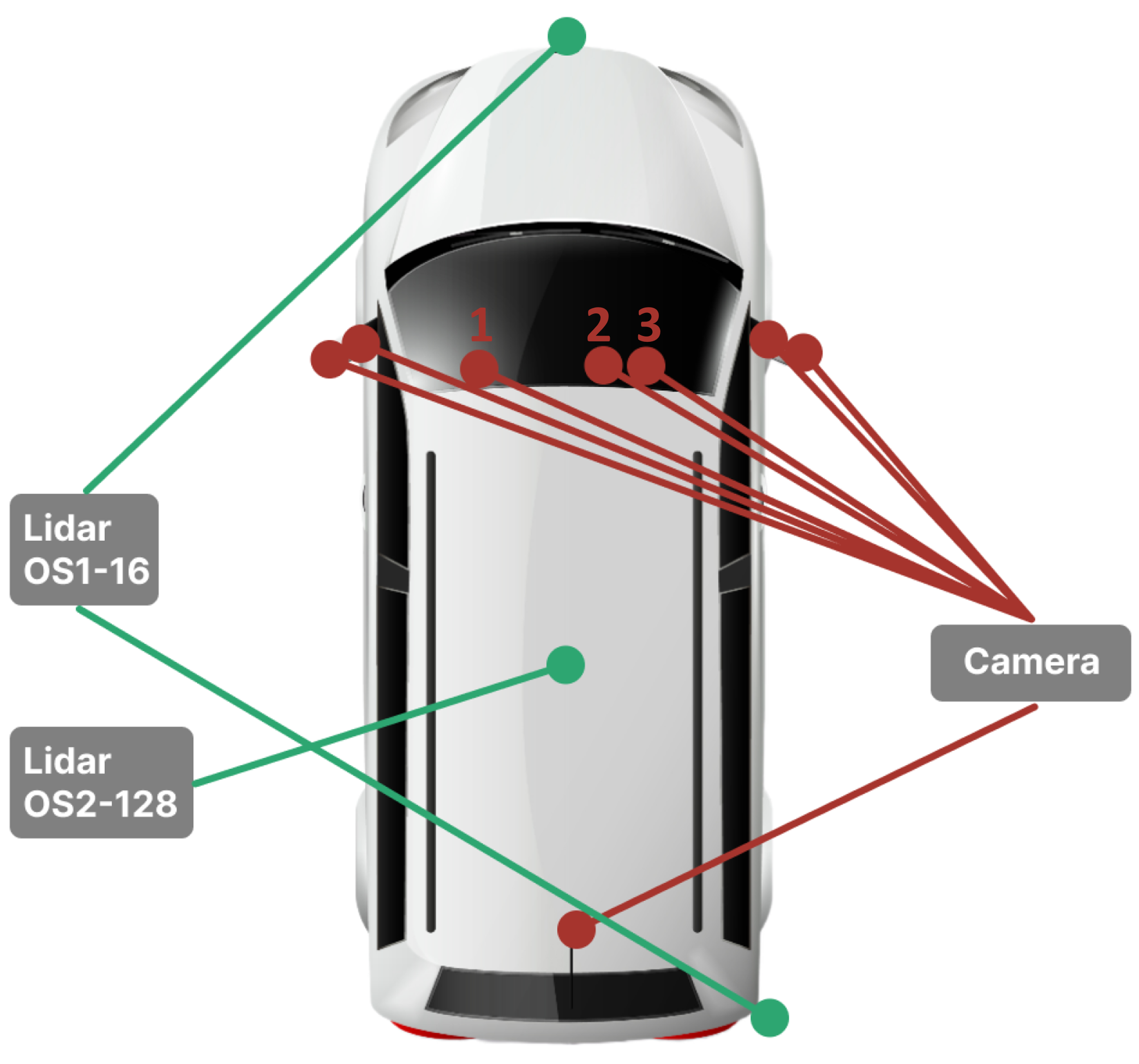}
    \caption{This figure illustrates the placement of the sensors on the autonomous vehicle at NAPLab, NTNU \cite{gusev2022remote}. }
    \label{fig:car2}
    
\end{figure}

The autonomous vehicle stationed at NAPLab has several sensors, including GNSS SwinftNav Duro Inertial with CPOS subscription, 8 cameras and 3 LiDARs. The placement of these sensors can be seen in Fig.\ref{fig:car2}. The three front-facing cameras inside the car's windshield were used for data collection. Cameras 1 and 2 are 60-degree FOV Sekonix SF3325 cameras, and camera 3 is a 120-degree FOV Sekonix SF3324 camera. All frames are 1920x1208 pixels. The car also has a drive-by-wire kit (DriveKit) CAN that enables access to the car's Diagnostic CAN interface, and car-related parameters such as speed and steering can be recorded. All the sensors record data using the same clock and are temporally synchronised. The sensors are interfaced with NVIDIA DRIVE AGX Xavier developer kit with DriveWorks 4.0.
\subsubsection{National Road Database (NVDB)} \label{sec:nvdb}
NVDB is a public database that contains information about most of the road objects in Norway, such as traffic signs, road networks, traffic accidents, etc. NVDB is maintained by the Norwegian Public Roads Administration and is used to implement and ensure good road quality. We used NVDB to collect positional data for traffic signs to serve as ground truth. Due to the inaccuracy and unreliability of the positional data of some traffic signs, this ground truth was discarded. We decided to collect our own ground truth by measuring the positional data using CPOS. However, it is worth noting that NVDB is continuously updated and rigorously maintained. The incorrect data can be reported and corrected. Therefore, data from NVDB can be used as ground truth. The code to pull data from the NVDB API will be provided on the GitHub repository for the paper.

\subsubsection{GNSS Positioning with CPOS Service} \label{sec:cpos}

\begin{figure}
\centering
     \begin{subfigure}[b]{0.49\textwidth}
         \centering
         \includegraphics[width=\textwidth, height=\textwidth, angle=-90, trim={1cm 1cm 1cm 1cm},clip]{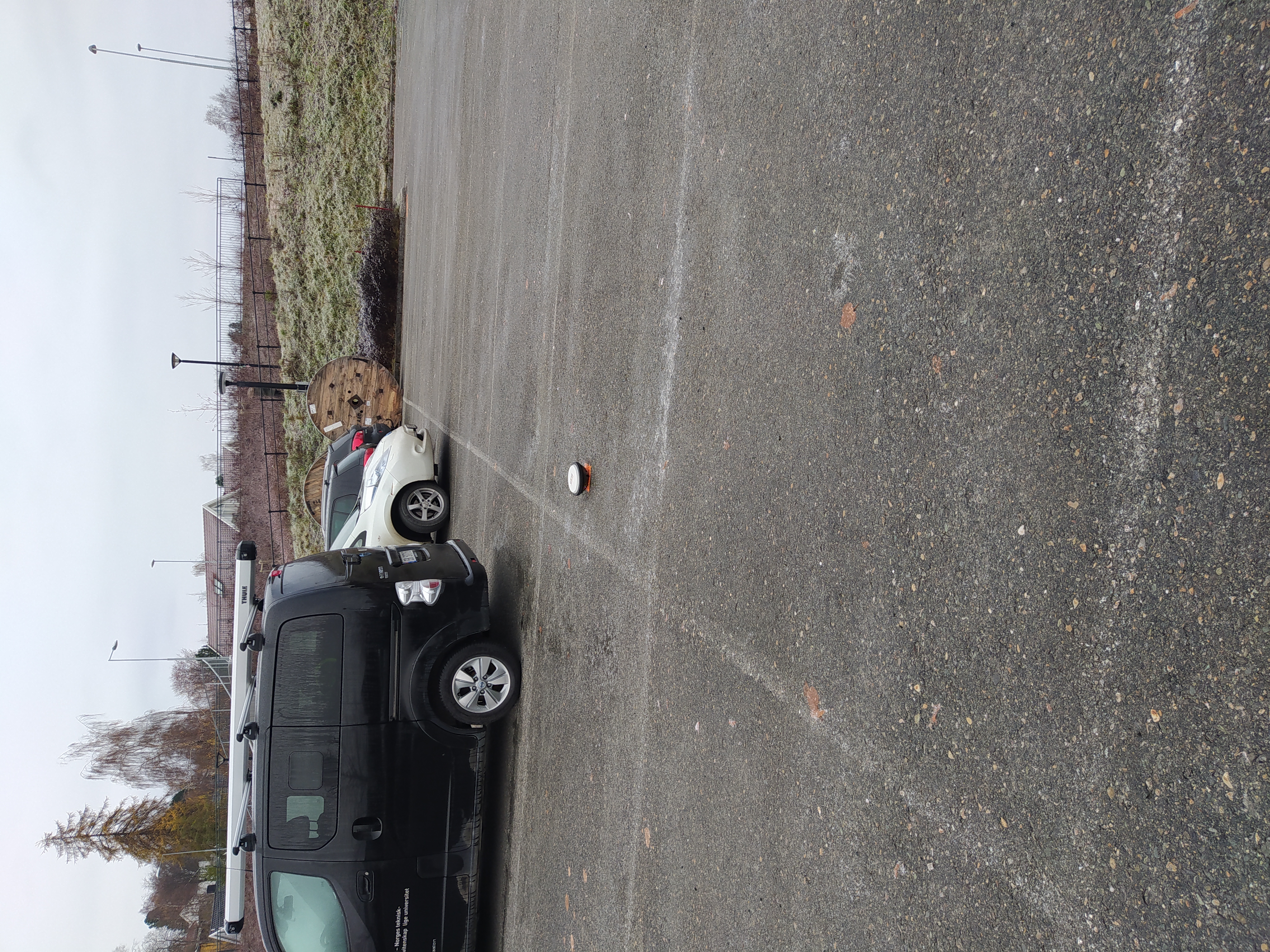}
         \label{fig:cpos}
     \end{subfigure}
     \begin{subfigure}[b]{0.49\textwidth}
         \centering
         \includegraphics[width=\textwidth, height=\textwidth, angle=-90]{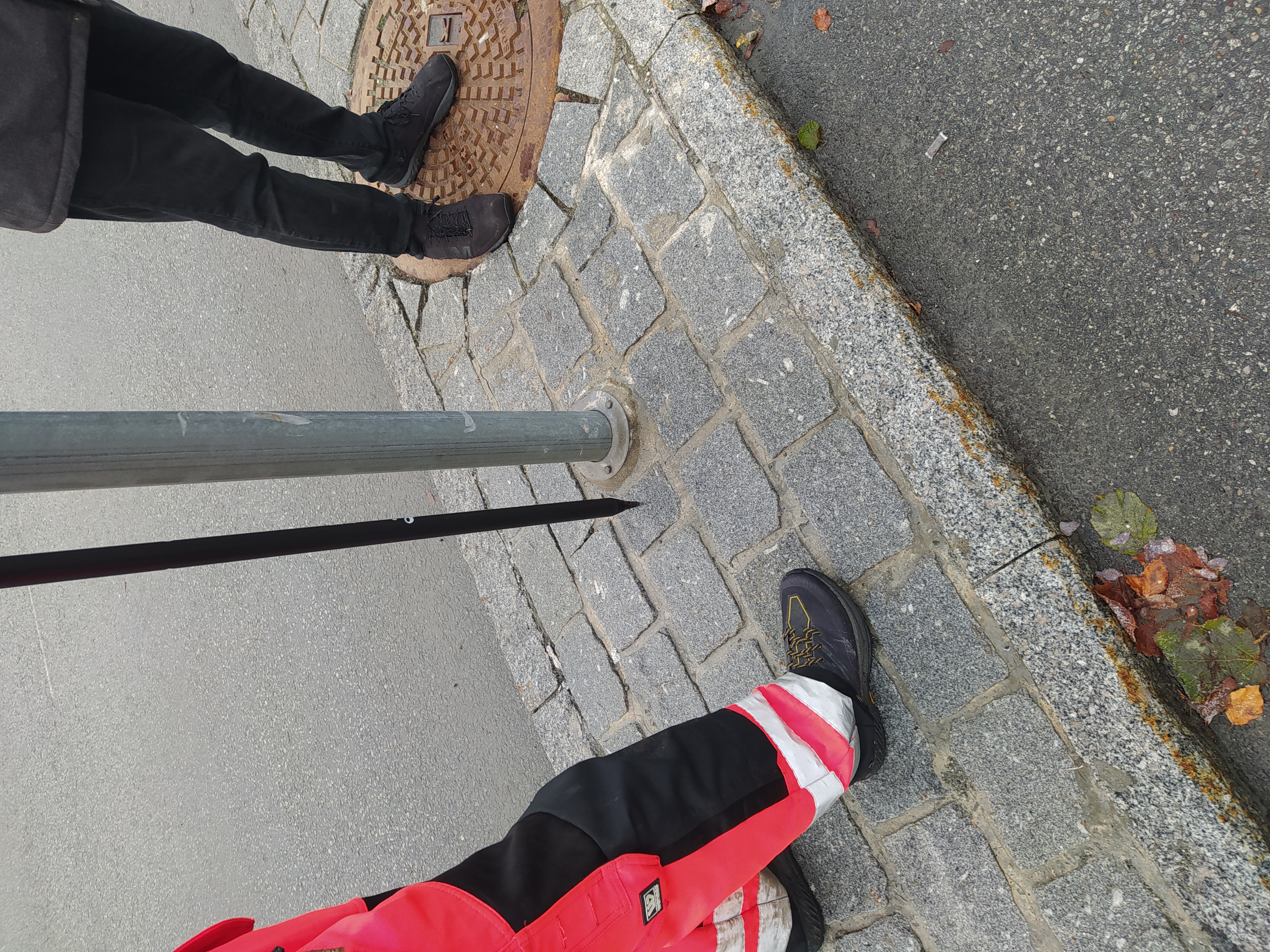}
         \label{fig:signcpos}
     \end{subfigure}
\caption{This figure illustrates the placement of the sensor for ground truth data. The left image shows the orange control marker in the NTNU Gloshaugen campus parking lot. The right image shows the measurement for traffic signs.} 
\label{fig:cposmeas}
\end{figure}

Since NVDB data cannot be a reliable ground truth, we gathered the geo-location data on several targets ourselves to validate our methodology. The GNSS receiver was used with the CPOS service enabled, which uses RTK Networks to measure positional data with centimetre accuracy. The ground truth for all the targets was measured using a standard device utilized by the road authorities. The sensor was placed in the centre of the orange control markers, while for traffic signs, the position of the point where the traffic pole meets the ground was measured as illustrated in Fig.\ref{fig:cposmeas}.  

\subsection{Camera Calibration using NVIDIA DriveWorks}
\label{sec:calib}
\begin{figure}
    \centering
    \includegraphics[scale=0.4]{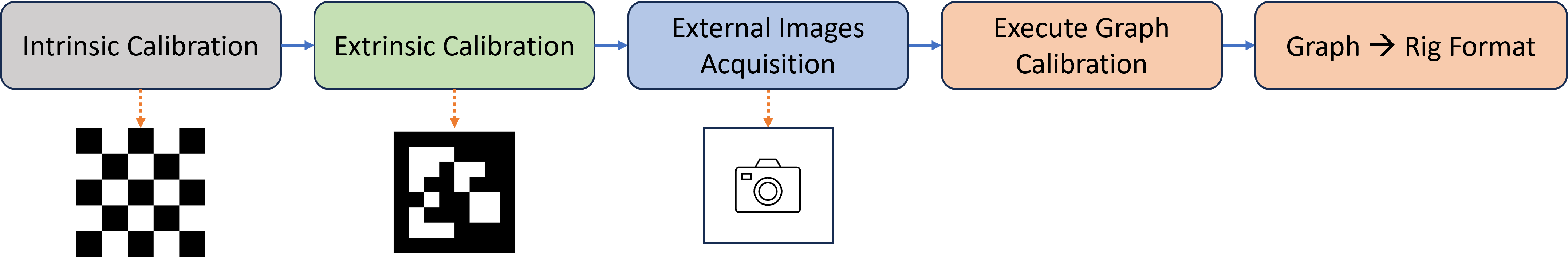}
    \caption{Flowchart for the camera calibration process using NVIDIA DriveWorks}
    \label{fig:calib_process}
\end{figure}

We used the autonomous vehicle at NAPLab, NTNU, to test our proposed method. The car runs on NVIDIA DRIVE OS 5.2.6 and NVIDIA DriveWorks 4.0. Using camera images, we calculate the target's distance from the vehicle with the functions provided by NVIDIA. This estimation is contingent on the accurate calibration of the cameras. As mentioned in NVIDIA's documentation for DriveWorks 4.0\cite{NvidiaCalib}, the calibration process was followed. Individual steps can be seen in the process diagram in Fig. \ref{fig:calib_process}. 

\begin{figure}
  \centering
  \begin{tabular}{ c }
    \includegraphics[width=0.99\linewidth,height=0.25\linewidth]{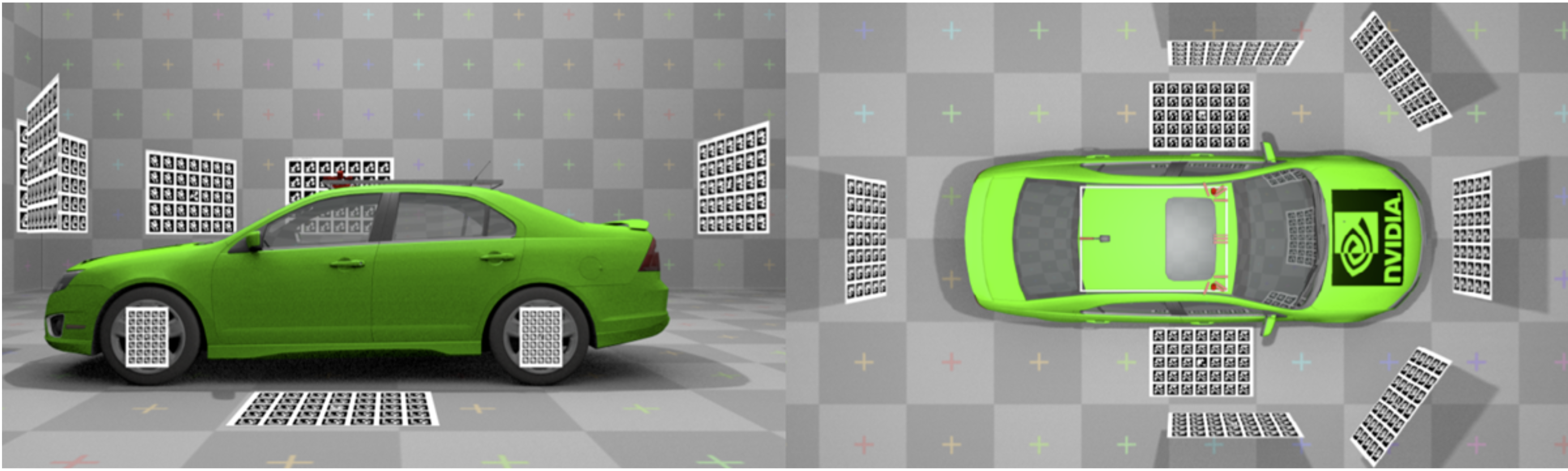} \\
    \includegraphics[width=.49\linewidth,height=0.2\linewidth]{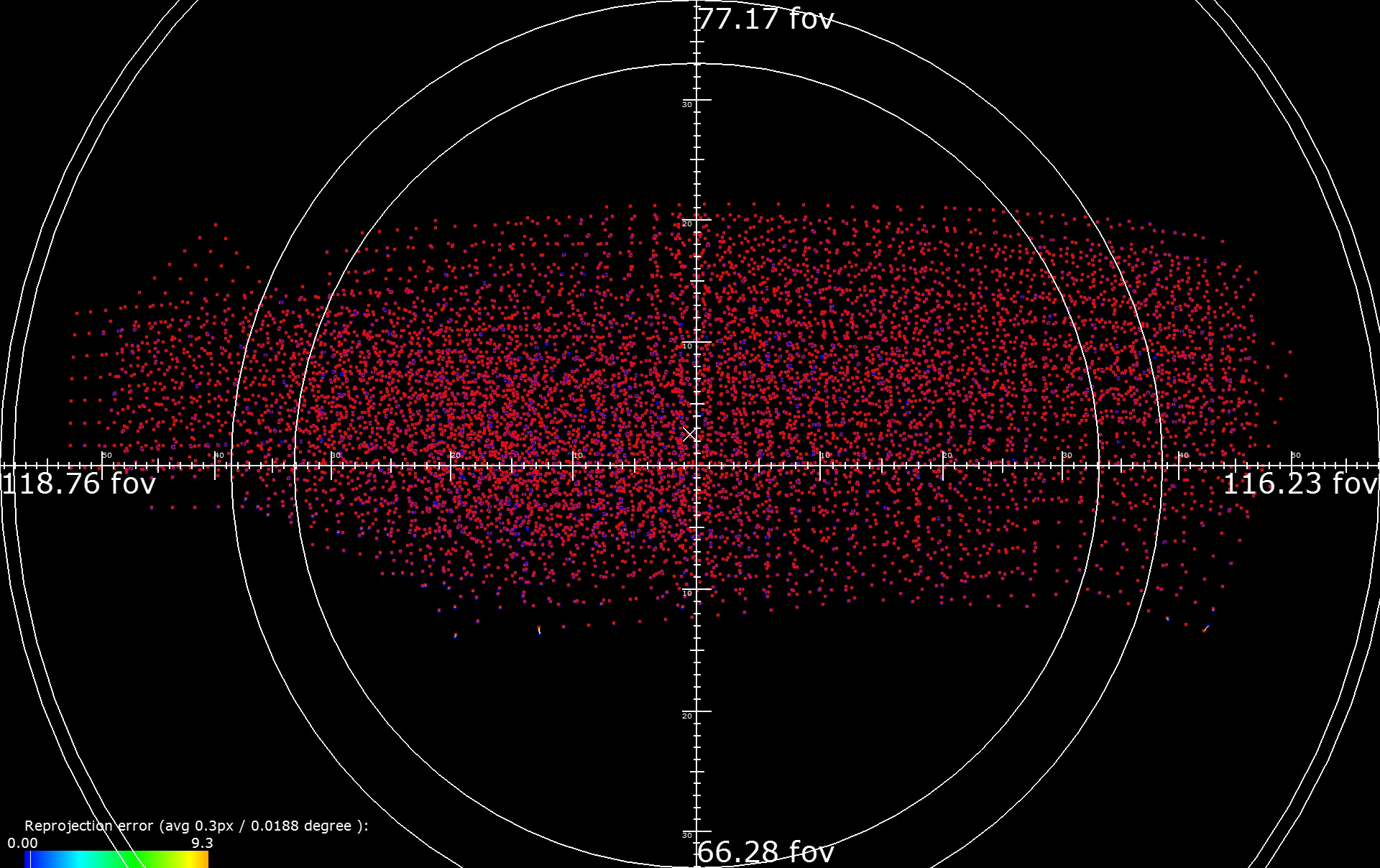} 
    \includegraphics[width=.49\linewidth,height=0.2\linewidth]{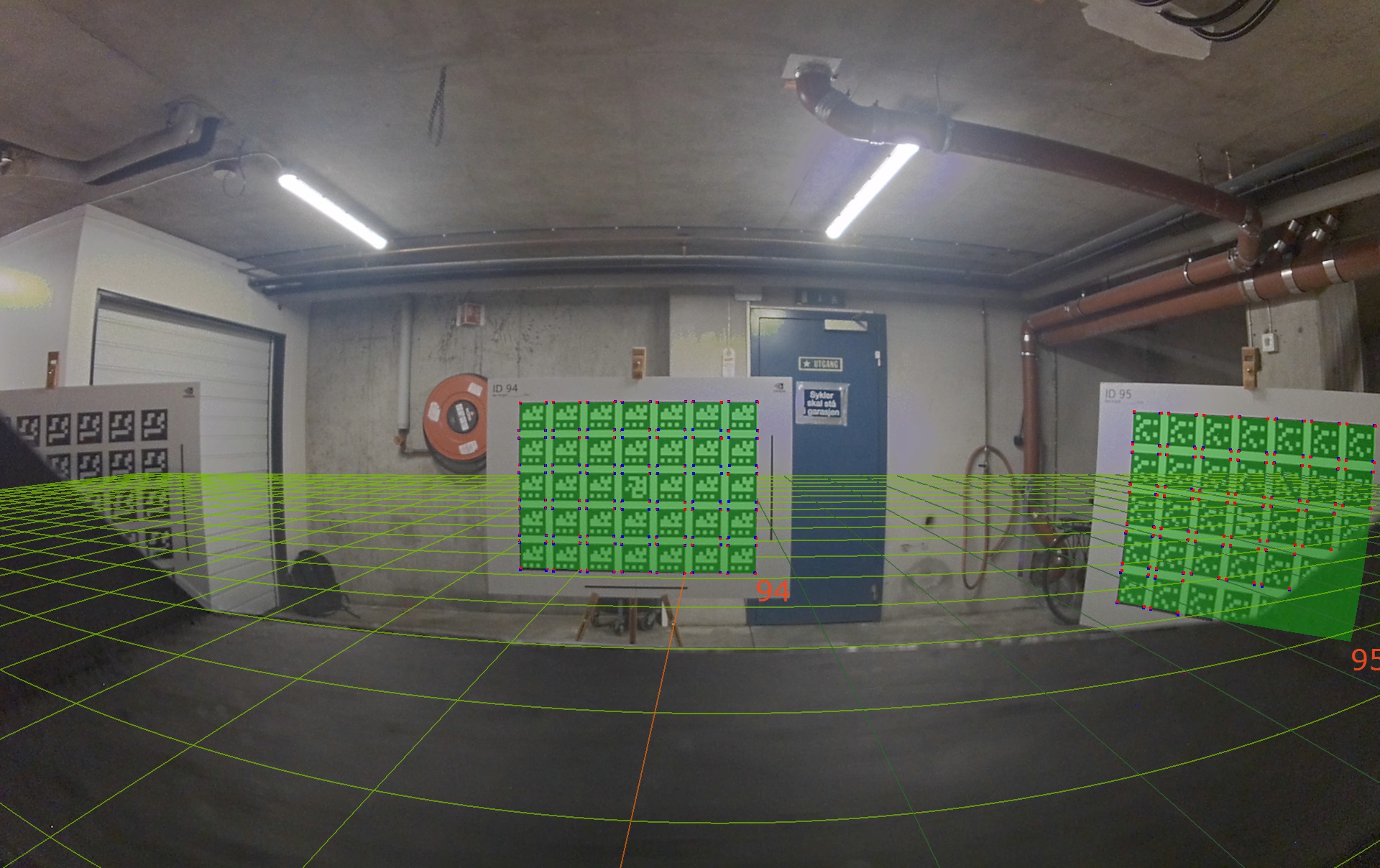} 

  \end{tabular}
  \caption{The top image shows the placement of the AprilTag targets around the car for extrinsic calibration. The image is taken from NVIDIA DriveWorks documentation\cite{NvidiaCalib}. The left image in the bottom row shows the validation of the intrinsic calibration of camera 3, and the right image validates the extrinsic calibration of camera 3. }
  \label{fig:calibration}
\end{figure}

NVIDIA provides three different types of calibration models: pinhole, OCAM\cite{mei2007single} and ftheta. We calibrated Camera 1 and Camera 2 using the OCAM model\cite{mei2007single} and Camera 3 using the ftheta model. We tested the ftheta model with Camera 1 and Camera 2 as well, but it did not give good results, leading us to believe that the ftheta model fits best with the cameras having a fish-eye effect. The details of the ftheta model are provided in a white paper, which is shared on the NVIDIA DriveWorks Forum\cite{ftheta}. NVIDIA claims that its implementation is the closest to Courban et al. \cite{courbon2007generic} but does not provide any quantitative analysis for the accuracy of their calibration. We verified the calibration by estimating the distance of an object placed at a known distance from the car. The calibration setup and the results from the intrinsic and extrinsic calibration can be seen in Fig. \ref{fig:calibration}. 
In addition to the guidelines provided by NVIDIA\cite{NvidiaCalib}, we state our own recommendations here, which we concluded after repeated calibration trials.
\begin{itemize}
    \item The checkerboard target must be used for intrinsic calibration, and the AprilTag target must be used for extrinsic calibration.
    \item The target must be entirely rigid for calibration to work.
    \item The car must be on a completely flat surface for extrinsic calibration. An incline of even 0.5 degrees may cause the calibration to fail.
    \item The ftheta calibration model is the most robust but is suited to cameras with fisheye effects.
    \item The stability threshold was increased to 2.0 for intrinsic calibration. The effect of this is uncertain as NVIDIA provides no documentation for the threshold parameters.
\end{itemize}

\subsection{Geolocation Estimation}
\label{sec:geoest}

\subsubsection{Estimating distance of the target from the camera on the vehicle:}
After the rigorous calibration of the onboard cameras, we employ NVIDIA DriveWorks for target distance estimation relative to the vehicle. Points of interest in the frames acquired by the cameras were manually annotated. With the pixel coordinates established, the corresponding optical ray emanating from the camera to that specific pixel can be deduced using the functions provided by the NVIDIA DriveWorks Framework. This ray vector's magnitude represents the target's depth or distance from the camera. It's imperative to account for the camera's projection matrix with respect to the vehicle's rig and the transformation from the vehicle rig to the world coordinate system. The code is written in cpp and will be provided on the GitHub repository for the paper.

\subsubsection{Estimating heading of the target from the car:}
The target's heading relative to the car can be calculated by using equation \ref{eq:heading}.

\begin{equation}
    \angle = \left( \frac{FOV_{camera}}w \right) \times \left( p_{x} - \frac w{2} \right) + \psi_{camera} + \theta_{car}
\label{eq:heading}
\end{equation}

Where $FOV_{camera}$ is the field of view of the camera, $w$ is the width of the image, $p_{x}$ is the pixel position of the target along the x-axis, $\psi_{camera}$ is the yaw angle of the camera and $\theta_{car}$ is the heading of the vehicle.

\subsubsection{Geolocating the target:}
We have the position of the vehicle specified by its latitude (\(\phi_1\)) and longitude (\(\lambda_1\)), and we want to find the latitude (\(\phi_2\)) and longitude (\(\lambda_2\)) of a point located at a known distance \(d\) with a heading \(\theta\) relative to the vehicle. Thus, we can use the inverse haversine formula given in the following equations.

\begin{equation}
 \phi_2 = \arcsin\left( \sin(\phi_1) \cdot \cos\left(\frac{d}{R}\right) + \cos(\phi_1) \cdot \sin\left(\frac{d}{R}\right) \cdot \cos(\theta) \right) 
\end{equation}

\begin{equation}
 \lambda_2 = \lambda_1 + \arctan2\left( \sin(\theta) \cdot \sin\left(\frac{d}{R}\right) \cdot \cos(\phi_1), \cos\left(\frac{d}{R}\right) - \sin(\phi_1) \cdot \sin(\phi_2) \right) 
\end{equation}

Where \(\phi_2\) is the target latitude, \(\lambda_2\) is the target longitude,\(d\) is the distance to the target, \(R\) is the Earth's radius, \(\theta\) is the heading (in radians) of the target from the vehicle.

\section{Experiments}
We have tested our method in different scenarios to prove its robustness and efficacy. The target(s) is always stationary, while the vehicle speed varies for different scenarios. The target, experimental setup and results for each scenario are discussed in the following text. We apply the same method to the control markers and traffic signs to prove that our method is general and applicable to all road objects that can be detected using a camera. Experiments related to the following scenarios were carried out.
\begin{enumerate}
    \item Vehicle is stationary in a controlled environment.
    \item Vehicle is moving towards the target in a controlled environment. 
    \item Driving in urban areas at lower speeds.
    \item Driving on a highway at higher speeds.
\end{enumerate}

\textbf{Coordinate Reference System:} Measurements from the autonomous vehicle were in WGS84 long/lat format. Measurements from the CPOS were in UTM33 format. All measurements were converted to WGS84 long/lat format for the calculations. The transform function from the Pyproj library was used to transform from UTM33 to WGS84, and the Geod function was used to calculate the error, i.e. the distance between ground truth and estimated position.

\section{Results and Discussion}

\subsection{Stationary Vehicle in a Controlled Environment}
In this scenario, we evaluated the performance of three front-facing cameras mounted on the stationary vehicle to estimate the distance of an orange control marker using our proposed method. The results are shown in Table \ref{tab:stat}. Camera 1 showed a significant deviation at the 19.3-meter mark. Camera 2 produced relatively lower errors, suggesting improved distance estimation accuracy, but still displayed a noticeable error at the longest distance. In contrast, Camera 3 exhibited the most promising results with minimal errors across the three distances, implying that this camera's measurements were most closely aligned with the actual distances. This difference is observed because Camera 3 was calibrated with a different model, as discussed before. Table \ref{tab:stat} shows that the error increases as the target is placed further away from the vehicle.

\begin{table}[!ht]
\caption{Estimated position error when the vehicle is stationary. All values are in meters.}\label{tab:stat}
\centering
\begin{tabular}{|c|c|c|c|}
    
\hline
Distance to the Target &  Camera 1 (60FOV, OCAM) & Camera 2 (60FOV, OCAM) & Camera 3 (120FOV, ftheta)\\
\hline
9.004 &  0.77 & 0.82 & 0.34\\
\hline
11.78 &  1.53 & 1.42 & 0.22\\
\hline
19.3 &  5.05 & 4.53 & 1.01\\
\hline
\end{tabular}

\end{table}

\subsection{Moving Vehicle in a Controlled Environment}

The experiment was set up such that the vehicle accelerated towards the target and stopped as the target moved out of the frame. The target was an orange control marker in the parking lot of the NTNU campus. We think of the parking lot as a controlled environment because it is an open space with no buildings or trees obstructing the GNSS signals.

\subsubsection{Effect of error vs distance}

\begin{figure}
  \centering

\includegraphics[width=0.5\textwidth]{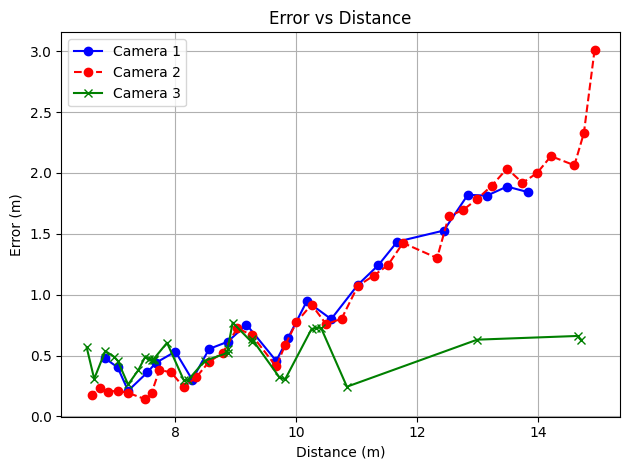}
\caption{Estimated position error when the vehicle moves towards the target. } \label{fig:mov}
\end{figure}

In the moving vehicle experiment, we also assessed the performance of the three front-facing cameras mounted on the moving vehicle that approached an orange control marker with acceleration. The car accelerates toward the target at 15 meters and stops at 6 meters from the target when it goes out of the camera frame. The measured error is illustrated in the graph in Fig.\ref{fig:mov}. Cameras 1 and 2 showed a fluctuating error pattern as the vehicle approached the marker. In contrast, Camera 3's errors were comparatively consistent and lower across the measured distances. Therefore, it can be inferred that Camera 3 offers the most reliable distance measurements in dynamic conditions using our inverse haversine formula-based method. Due to the results of this experiment, we used Camera 3 for the remainder of the experiments.

\subsubsection{Effect of Error vs Speed}

\begin{figure}
  \centering

\includegraphics[width=0.5\textwidth]{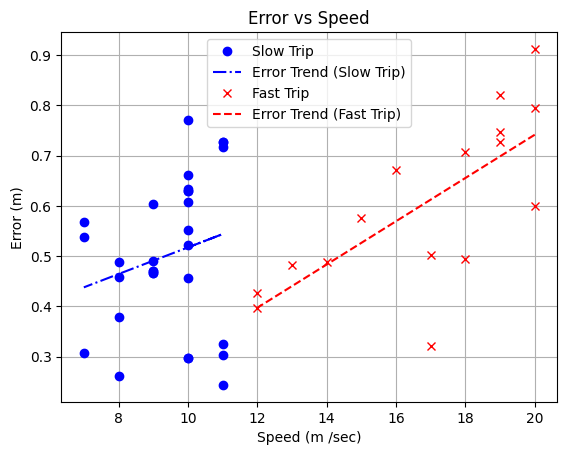}
\caption{Estimated position error when the vehicle moves towards the target at different speeds. } \label{fig:errorvspeed}
\end{figure}

In the subsequent phase of our experiment, we focused exclusively on Camera 3 to evaluate its performance under variable speeds, as shown in Fig. \ref{fig:errorvspeed}. With the same experimental setup as before, two distinct experiments were conducted: one at relatively slow speeds ranging from 7 to 11 km/h and the other at faster speeds ranging from 12 to 20 km/h. In the slow-speed experiment, the errors observed were generally consistent. On the other hand, during the fast-speed experiment, as expected, the error was generally higher. 

These findings indicate that while Camera 3 shows commendable performance in slow and fast-moving conditions, its error rate tends to be slightly elevated at higher speeds. Nonetheless, the variations in error do not show a simple linear relationship with speed, suggesting other influencing factors might be at play. The fluctuations can be thought of as a direct result of the acceleration and deceleration of the vehicle. They may occur due to the communication delay from the GNSS. The GNSS signal has a frequency of 5Hz while the camera frame rate is 30FPS. This delay between sensors contributes to the error, and its effect is enhanced while the vehicle accelerates or decelerates. This nuanced understanding of Camera 3's performance at different speeds provides valuable insights for applications requiring precise distance measurements under variable speed conditions using our methodology.

\subsection{Driving in an Urban Area}
\begin{figure}
  \centering
  \begin{tabular}{ c }
    \includegraphics[width=.49\linewidth,height=0.4\linewidth]{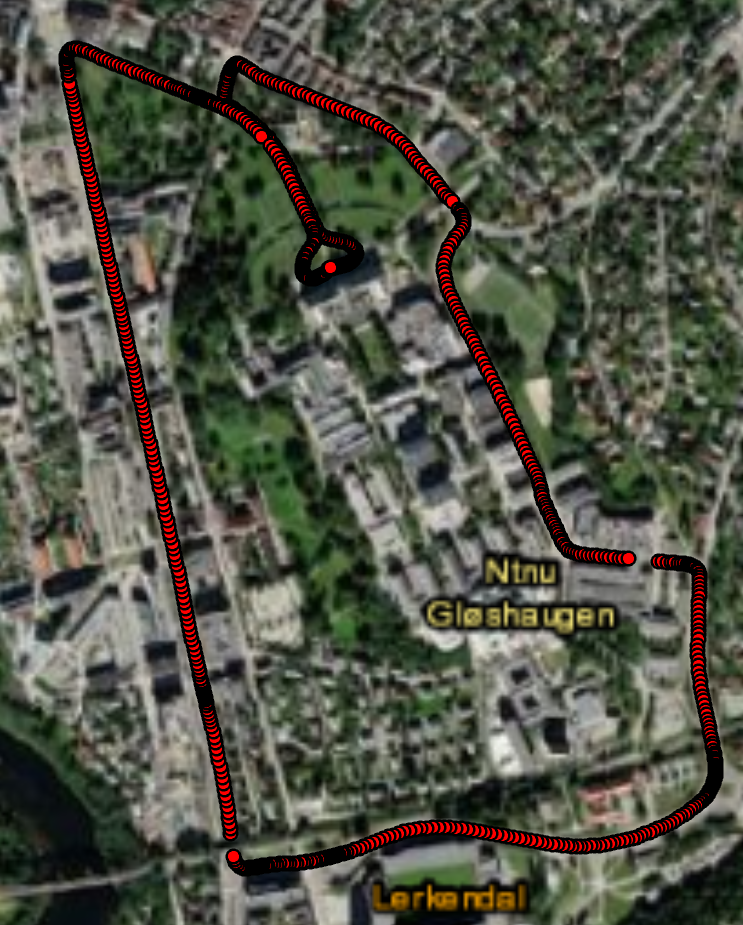}
  \end{tabular}%
  \begin{tabular}{ c }
    \includegraphics[width=.49\linewidth,height=0.2\linewidth]{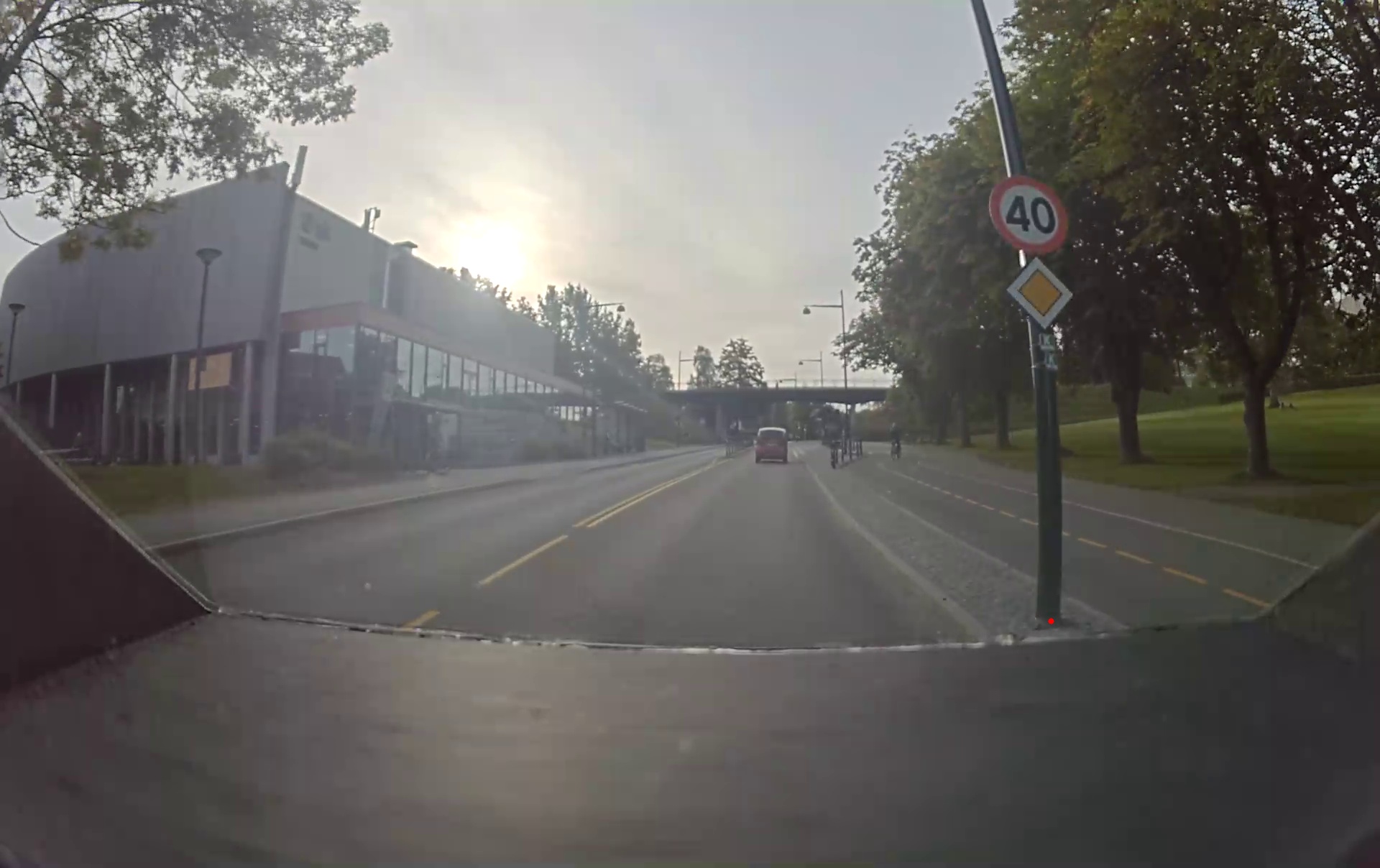} \\
      \includegraphics[width=.49\linewidth,height=0.2\linewidth]{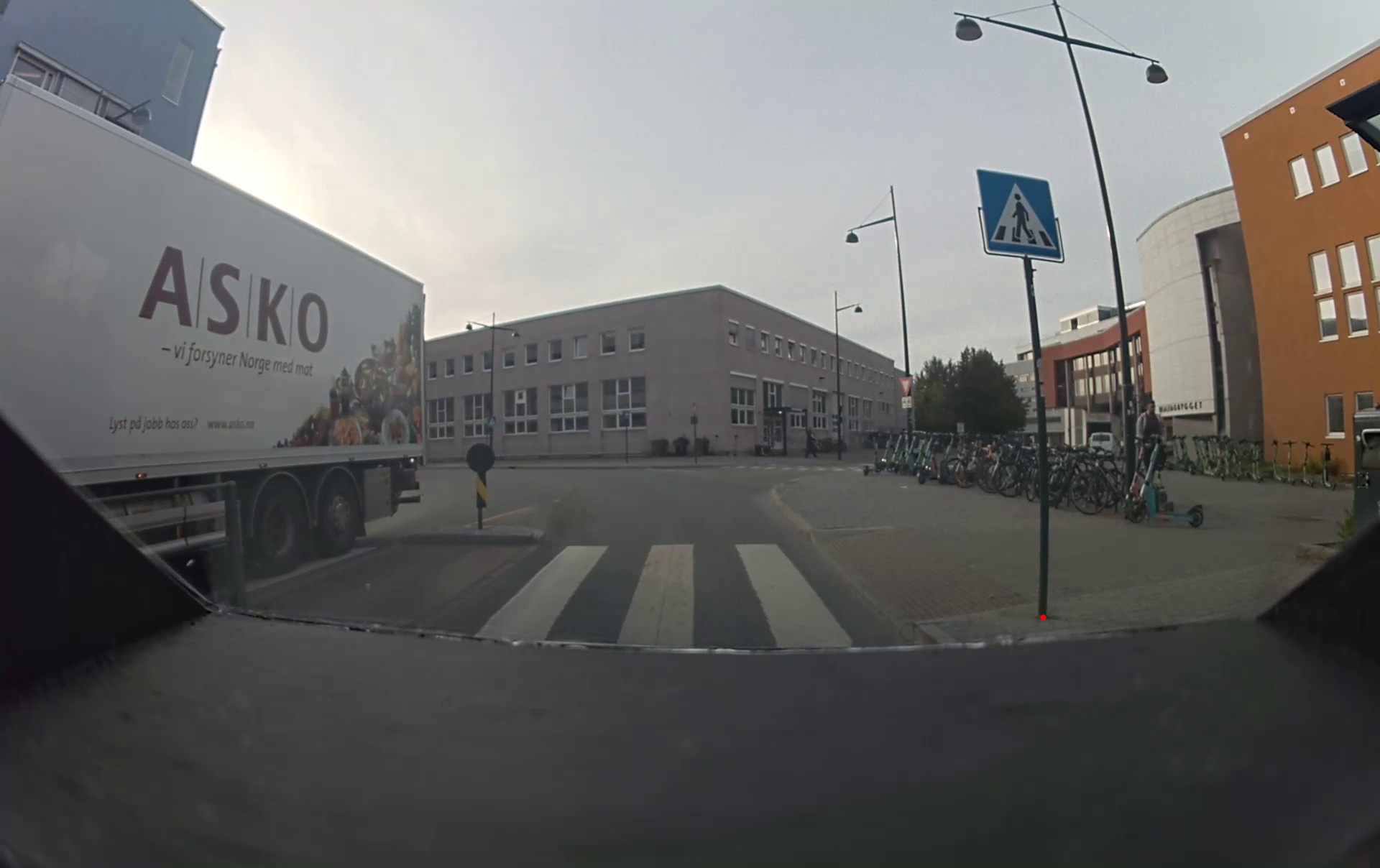} \\

  \end{tabular}
  \caption{The left image shows the route around the NTNU campus where the experiment was done. The right column shows images from Camera 3, where the bottom of the signs is labelled for geolocation.}
  \label{fig:glos_images}
\end{figure}

We evaluated the performance of Camera 3 in an urban environment, specifically the NTNU campus in Trondheim. The route is shown in Fig.~ \ref{fig:glos_images}. This was done to gauge how the camera would fare outside controlled conditions where buildings, trees or bridges sometimes obstruct the GNSS signal. 13 traffic signs were chosen as targets. 

\begin{figure}
\centering
     \includegraphics[scale=0.35]{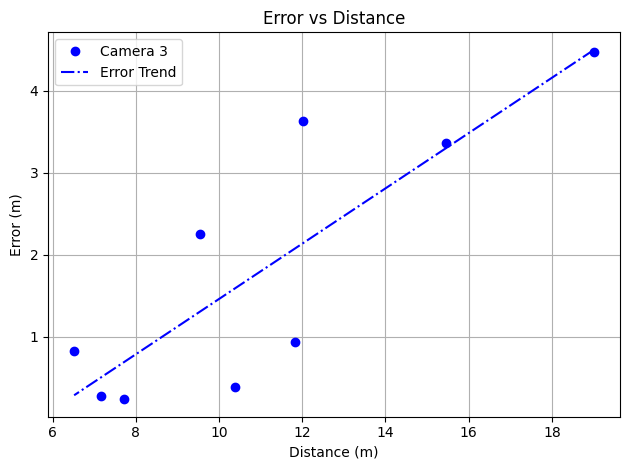}
     \label{fig:glos1}
\caption{Estimated position error when driving around the NTNU campus. The dotted line shows the error trend with an increase in distance.} 
\label{fig:glos}
\end{figure}

As shown in Fig. \ref{fig:glos}, the error rates for this test ranged from as low as 0.26 meters at a distance of 6.9 meters to as high as 3.9 meters at 14.6 meters.  This suggests that the camera is subject to several external factors in the urban setting, such as lighting conditions, obstructions, or other environmental variables that may impact its accuracy adversely. These results indicate that while Camera 3 maintains a level of reliability, its performance can be significantly influenced by external factors in an urban environment. 

\subsection{Driving on a National Highway}

\begin{figure}
  \centering
  \begin{tabular}{ c }
    \includegraphics[width=0.99\linewidth,height=0.13\linewidth]{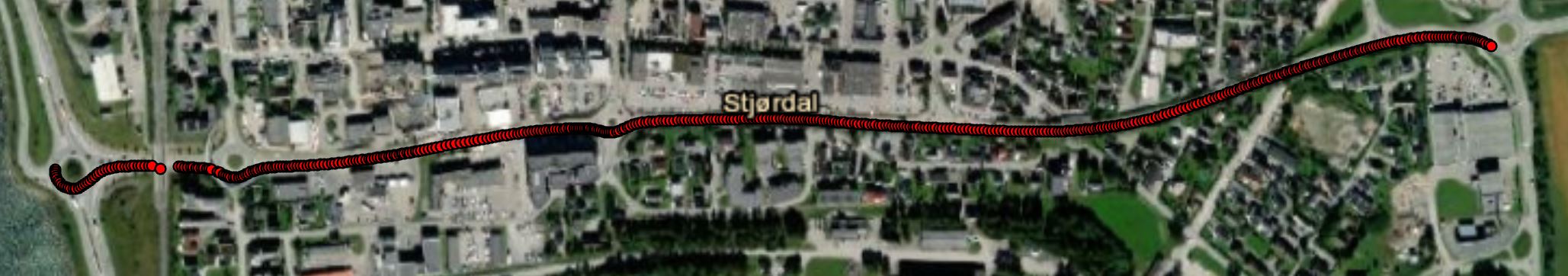} \\
    \includegraphics[width=.49\linewidth,height=0.2\linewidth]{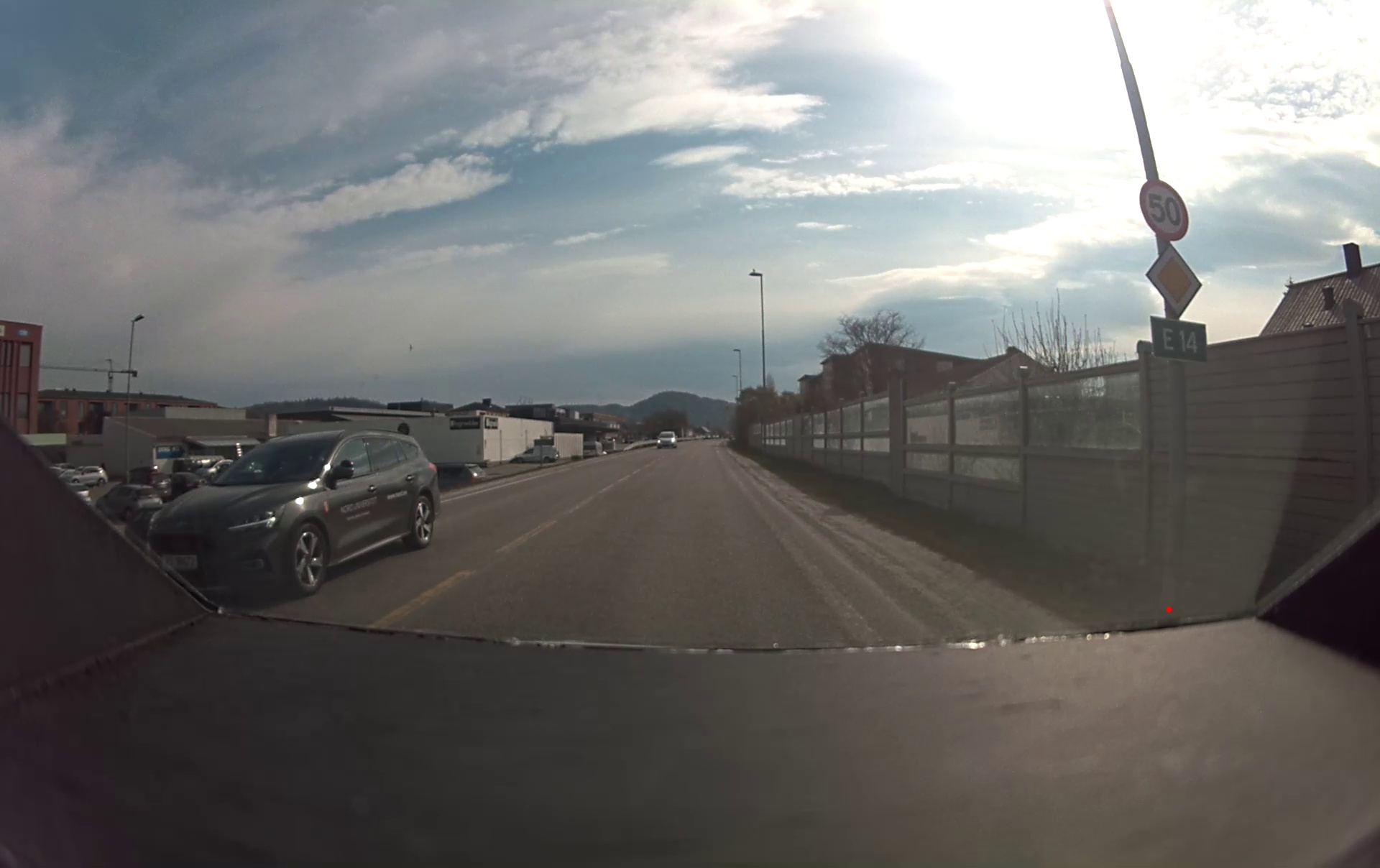} 
    \includegraphics[width=.49\linewidth,height=0.2\linewidth]{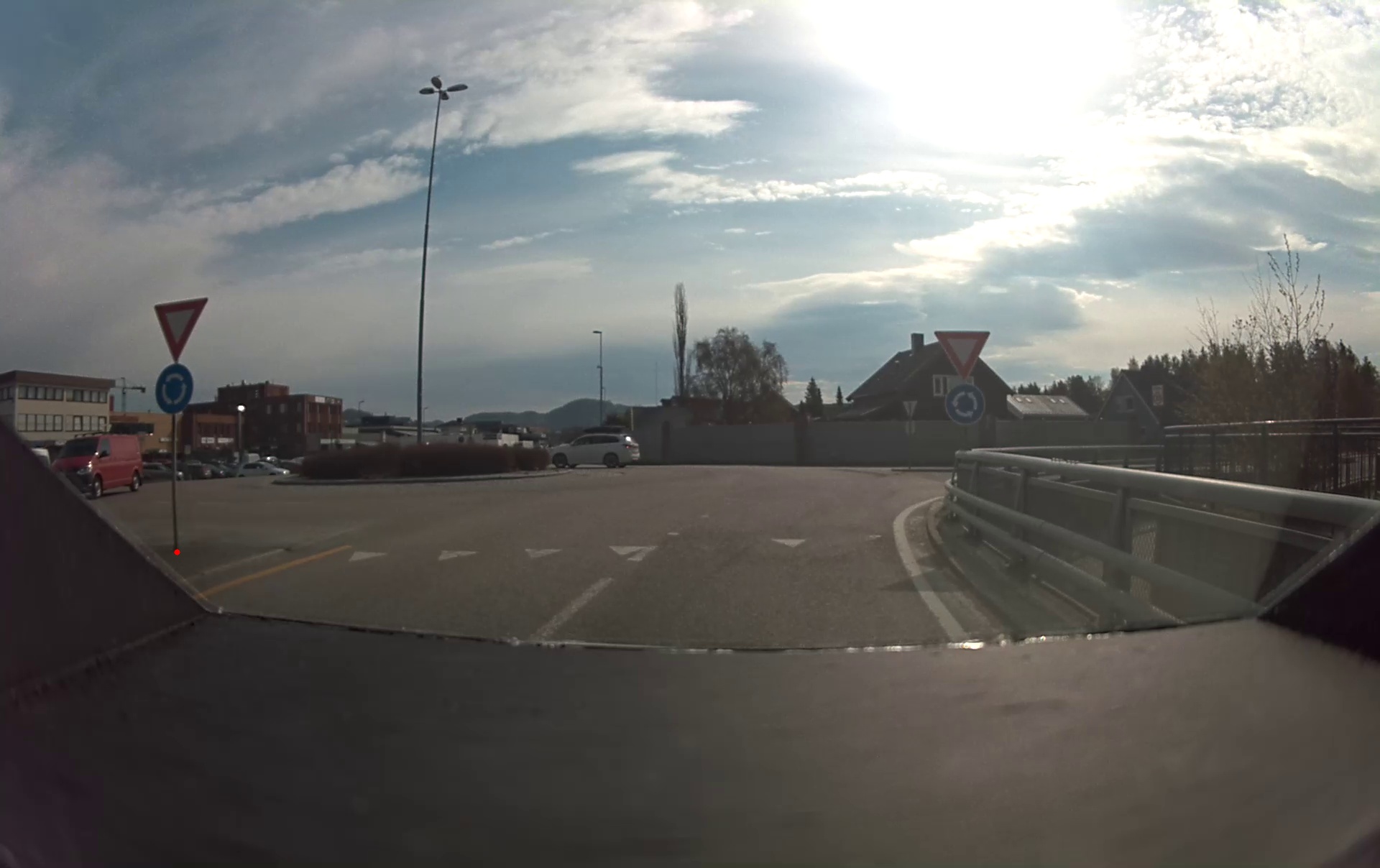} 

  \end{tabular}
  \caption{The top image shows the route on EV14 where the experiment was done. The bottom row shows images from Camera 3, where the bottom of the signs on EV14 are labelled for geolocation.}
  \label{fig:ev_images}
\end{figure}

In addition, we evaluated the performance of Camera 3 under dynamic highway conditions. Fig.\ref{fig:ev_images}  shows the section of EV14, which was used as the test area. 16 traffic signs were chosen as targets. 

\begin{figure}
\centering
     \includegraphics[scale=0.35]{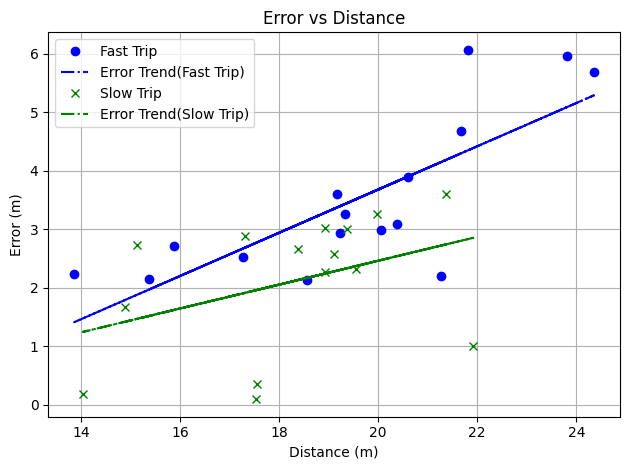}
\caption{Estimated position error when driving on EV 14. The dotted lines show that the error increases at higher speeds.  } 
\label{fig:ev14}
\end{figure}

Fig. \ref{fig:ev14} shows the resulting errors in the estimated positions of these 16 traffic signs versus the target's distance from the vehicle. During the slow-speed trial, the speed was an average of 35 km/h; during the fast-speed trial, the speed was 45 km/h. Intriguingly, increased speed did not universally correspond to higher error rates, but the errors were generally inconsistent across a range of distances in both speed categories. This inconsistency underlines the complexity of the camera's performance in a dynamic highway environment and suggests that external factors, such as vehicle speed, road conditions, sensor communication delays, or other variables, significantly influence its measurement accuracy. However, the general trends observed are as expected. The error increases at higher speeds as the target moves further from the vehicle. We can restrict the measurement radius of the targets from the car. The typical measurement radius of commercially available equipment is 10 meters. If we set our measurement radius to 15 meters, we can still conclude that our method can geolocate targets with an error of up to 4 meters at high speeds.

\section{Conclusion and Future Works}

In this work, we utilized the NVIDIA DRIVE platform integrated with a monocular camera system and CPOS for an accurate and real-time geolocalization of road objects. We share an effective method to ensure successful calibration of the monocular camera setup on the NVIDIA DriveWorks framework. Extensive experiments demonstrate that our approach maintains a low error in all complex scenarios. 

This work is a precursor to a larger project in which the goal is to build a digital twin of the road network of Norway for more efficient maintenance for the Norwegian Public Roads Administration (NPRA). The proposed method will be improved in several aspects before being integrated with the digital twin. In the future, we will integrate this system with a neural network object detector to ensure automated annotation of road objects instead of manual annotations. Accounting for the communication delays from the sensors can also reduce measurement errors. We will also use all three cameras instead of just one to estimate an average location and minimise error. The methodology will be applied to detecting and geolocating road damages and other road objects.

\section*{Ethical Considerations}
\textit{All personally identifiable information, including pedestrians and vehicular registration plates, was blurred due to privacy protocols and ethical considerations.}

\section*{Acknowledgements}
\textit{This work was funded by the Norwegian Public Roads Administration (NPRA) as a part of the Smarter Maintenance project. The authors would like to thank Doreen Sibert and Dagfin Gryteselv (NPRA) for their valuable input to the project.}

\bibliographystyle{splncs04}
\bibliography{parts/bib}

\end{document}